\documentclass[letterpaper, 10 pt, conference]{ieeeconf}  

\IEEEoverridecommandlockouts

\usepackage{cite}
\usepackage{amsmath,amssymb,amsfonts}
\usepackage{algorithmic}
\usepackage{graphicx}
\usepackage{textcomp}
\usepackage{xcolor}
\usepackage[table]{xcolor}

\usepackage{amssymb}
\usepackage{siunitx}
\def\BibTeX{{\rm B\kern-.05em{\sc i\kern-.025em b}\kern-.08em
    T\kern-.1667em\lower.7ex\hbox{E}\kern-.125emX}}

\usepackage[final]{changes}

\newcount\Comments 
\usepackage{color}
\newcommand{\kibitz}[2]{\ifnum\Comments=1\textcolor{#1}{#2}\fi}

\title{FallGown: A Soft Pneumatic Wearable Device \\ for Fall Prevention
 \\

\author{Megan C. Coram$^{1}$, Haoran Lei$^{2}$, Steven H. Collins$^{1}$,\\ Steve B. Cousins$^{1}$, Michelle Baldonado$^{1}$, Cynthia Sung$^{2}$, and Allison M. Okamura$^{1}$}

\thanks{*This work was supported by the Stanford Robotics Center.}

\thanks{$^{1}$MC, SC, SC, MB, and AO are with the Department of Mechanical Engineering and Robotics Center, Stanford University, Stanford, CA, 94305, USA. e-mail: {\tt\small aokamura@stanford.edu}}%
\thanks{$^{2}$HL and CS are with the General Robotics, Automation, Sensing \& Perception (GRASP) Lab, University of Pennsylvania, Philadelphia, PA 19104, USA. e-mail: {\tt\small crsung@seas.upenn.edu}}%
}

\begin{document}

\maketitle
\thispagestyle{empty}
\pagestyle{empty}

\begin{abstract}
Concerns about falling and potential resulting injuries prevent many older adults from participating in forms of health-improving exercise such as walking, dance, and tai chi. Existing passive devices like walkers require use of the arms for support, and active devices like exoskeletons and robotic frames are often rigid, bulky, and stigmatized, which limits their use. The FallGown is a soft wearable support structure consisting of pressurized pneumatic pouches within a fabric shell; it can support the full weight of a user's body and aims to prevent falls. It is portable (6.9\,kg and 0.051\,m$^\mathbf{3}$ when stored), minimally burdensome (7.24-20.8\,N drag force depending on floor type and whether an optional rolling base frame is used), and convenient (less than 2.5 minutes on average to don and less than 0.5 minutes on average to doff). The optional rolling base lowers friction with the floor, but reduces portability and skirt deformation (which can be useful to fit through doors and other tight spaces). Fall prevention testing with a single user revealed a tradeoff between device size and stability against tipping; a skirt with 12 pneumatic pouches and a floor diameter of 1.4 meters with a base frame prevented all falls except in one case with a mechanical failure. On-board pressure sensors can sense the user's weight distribution, providing an opportunity for monitored exercise and adaptive support. 
\end{abstract}

\section{Introduction}

Wheelchairs and walkers are traditional passive devices that support mobility for adults with weakness, poor balance, and other movement challenges. Robotic versions include autonomous wheelchairs \cite{Gupta2026Evolution}, frames \cite{Bolli2025Elderly}, and handles \cite{Bolli2023Handle}. Exoskeletons reduce footprint and provide more versatile locomotor assistance in unstructured environments \cite{Siviy2023Opportunities}. However, the rigidity of traditional robotic mechanisms can create discomfort, and their bulkiness can limit portability. In a recent review of the state-of-the-art of robotics to assist mobility, physiotherapists and patients highlighted the following needs: safety, a one-size-fits-all approach, ease of device use, weight and placement of the device, cost of the device, and appearance of the device \cite{Morris2023State}. Negative perceptions of the appearance of an assistive device can result in stigma and reduce adoption, even for occupational scenarios \cite{Siedl2025Role}. Another recent survey found that fall prevention was the most (according to clinicians) or second-most (according to older adults) important feature that a wearable assistive device could offer \cite{Raitor2025}. 

Approaches using techniques from the field of soft robotics aim to make exoskeletons easier to wear and use, and potentially less noticeable. However, they can still be difficult to don and doff, and they are usually not able to support the user's full body weight. Exosuits are one form of “soft” support; they have no rigid frame to bear compressive loads, rather they use fabrics and tensile elements, usually cables, anchored to the body to apply forces in parallel with the wearer's own muscles \cite{Panizzolo2016Multijoint, SanchezVillamanan2019Compliant, Xiloyannis2022SoftSuits}. Some exosuits pass applied loads through the user's own joints, which could lead to painful changes in the magnitude and direction of joint loading.  Inflatable support devices (sometimes referred to as exomuscles) have been designed to support a single joint in research projects \cite{Simpson2017Exomuscle, Simpson2020Upper, Zhou2024InflatableShoulder} and commercial products like the torsional air-driven bladder used for knee support by Roam Robotics, Inc. The authors are not aware of any portable inflatable lower-body device that attempts full body-weight support.

\begin{figure}[t!]
    \centering
    \includegraphics[width=0.98\linewidth]{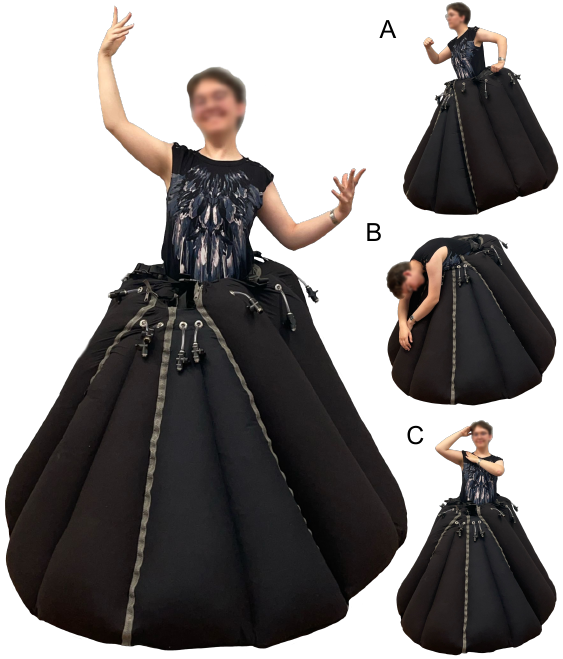}
    \caption{The FallGown is a soft pneumatic wearable device that can provide balance support and resist falls while (A) walking, (B) leaning, and (C) dancing.}
    \label{fig:fallgown}
\end{figure}

We were inspired to create a new form of mobility support device based on the following challenge presented to our research team by a local older adult day care center: During individual or group low-impact exercises, such as dance and tai chi, older adults' fear of falling will often cause them to opt out of physical activities or, if they choose to participate, keep them in a wheelchair or grasping a walker. Avoiding activity altogether for fear of falling can have negative impacts on long-term health \cite{Maki1997}, but reliance on current assistive devices prevents users from performing whole body movements, which are important to maintain strength and coordination and key to the aesthetics of the performance. Hands-free walkers such as the LifeGlider and Merry Walker permit arm movement, but they may still encounter the same stigma as traditional walkers, which limits their adoption. To address this challenge, we aimed to combine features of existing passive mobility support devices with soft wearable robotics. We created FallGown (Fig. \ref{fig:fallgown}), an inflatable skirt that is donned via a conventional climbing harness and zippers, can support a user's weight, and has a conical shape wide enough to resist falls for a variety of user movements. As a fashion statement, FallGown evokes a ball gown and is very noticeable; it counters the clinical aesthetic of prior assistive device designs.

The primary contribution of this work is a new inflated assistive wearable device that aims to support hands-free exercise for people with weakness, poor balance, and other movement challenges that limit beneficial exercise. A secondary contribution is a series of evaluation methods that address key assistive technology needs identified by patients and physiotherapists.

\section{Design and Fabrication}

The FallGown design is motivated by the need for a stand-alone device that is portable and comfortable, leaves the arms free, allows walking and knee bends, prevents falls, can adjust its level of support depending on user needs, and invokes fashion imagery rather than resembling a medical device. The primary design concept is that an inflated structure can provide body support while minimizing device weight. For portability, we constructed the device from thin, pliable textiles so that it can be easily folded up when deflated. For user comfort, we used soft cotton for the outermost layer of the device and made the harness adjustable. To free the arms, we designed the device to surround and support the body in all directions. To permit internal movement and promote stability, we made the base of the structure wide. For adjustable support and robustness, we made each pneumatic pouch individually inflatable. For an appealing appearance and durability, we top stitched the fabric seams.

\begin{figure}[t]
    \centering
    \includegraphics[width=0.98\linewidth]{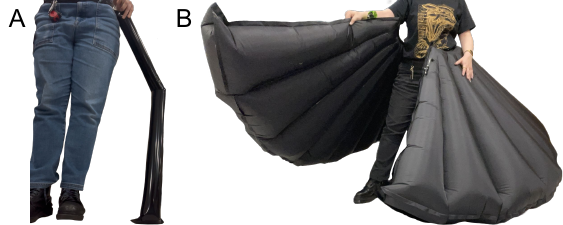}
    \caption{Initial FallGown design concepts that motivated the final design presented in Figs.~\ref{fig:fallgown} and \ref{fig:components}. (A) A single, narrow inflated tube buckles under the load of a user leaning on it, motivating the need for a continuous surface with robust structural design. (B) An initial FallGown prototype with a contiguous inflated body made of one layer of thermoplastic polyurethane (TPU)-coated ripstop nylon. The drawback of this approach was catastrophic failure in the cases of fabrication defects and puncture.}
    \label{fig:process}
\end{figure}

Initial development investigated the potential of using discrete pressurized tubes to support the body. However, these thin tubes buckled too easily outwards, as shown in Fig.~\ref{fig:process}A. Connecting the tubes together side-by-side resulted in the hollow cone shape. When compressed, the geometry is maintained by hoop stress. The first full-size prototype of the FallGown, pictured in Fig.~\ref{fig:process}B, was made from a single layer of thermoplastic polyurethane (TPU)-coated ripstop nylon in order to make the device as lightweight as possible. However, complex corner seams caused wrinkling issues in the ultrasonic welding process and made airtightness difficult to achieve. The first version of the skirt was comprised of one contiguous air pocket, so small holes or tears anywhere on the device would cause the whole structure to deflate. To address these issues, the final version of the device was composed of two layers: simple discrete rectangular pneumatic pouches on the inside and a robust protective fabric shell on the outside. The pouches are airtight and create individually strong members, and the fabric shell creates the overall structure and prevents buckling of the individual members.

\begin{figure*}[!h]
    \centering
    \includegraphics[width=0.98\textwidth]{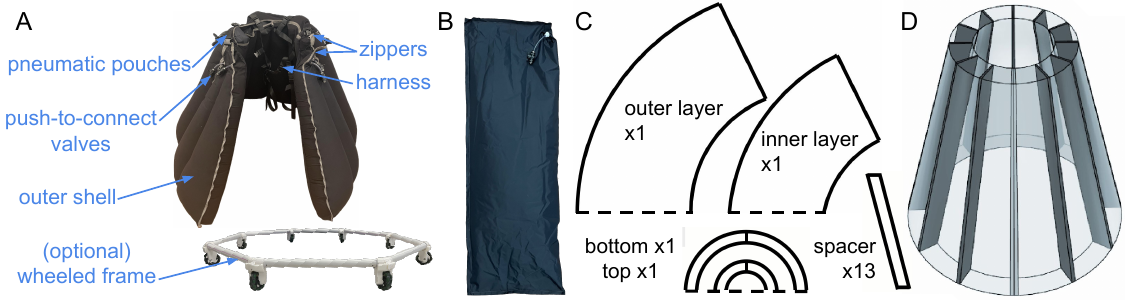}
    \caption{(A) The components of the FallGown include pneumatic pouches, push-to-connect valves, an outer fabric shell, accessible zippers, a waist harness with leg straps, and an optional wheeled PVC frame. (B) A single pneumatic pouch is a folded flat sheet of 70 denier ripstop nylon coated with thermoplastic polyurethane that is ultrasonically welded on the three non-folded edges. A tube, valve, and push-to-connect bulkhead connector are placed at the top right corner for inflation/deflation access. The pouch shown here is uninflated. (C) The sewing pattern for the outer shell consists of the outer cone, the inner cone, the top ring, the bottom ring, and 13 inner spacers. Solid lines indicate cuts, while dotted lines indicate mirroring. Seam allowance must also be added to the patterns according to the sewist's preference. (D) A 3D rendering of the fabric shell shows how the inner spacers are evenly distributed around the device. In the front of the skirt, two spacer panels are exposed on the exterior surface to create the zippered opening used to don and doff the device.}
    \label{fig:components}
\end{figure*}

The components of the final version of the FallGown evaluated in the following section are laid out in Fig.~\ref{fig:components}A. The clothing pattern shown in Fig.~\ref{fig:components}C is sewn together into the 3-dimensional form illustrated in Fig.~\ref{fig:components}D with two concentric cones linked axially along vertical lines. The fabric shell is made from lightweight cotton (AK Trading Co. Medium Weight Muslin), while the inner pouches are made from ultrasonically welded TPU-coated ripstop nylon (Quest Outfitters \#1056). Each pouch, shown in Fig.~\ref{fig:components}B, is inflated through a push-to-connect bulkhead connector and valve. Each pouch measures 145\,cm by 51\,cm when uninflated, which is oversized compared to the dimensions of the shell to ensure that the pouches can expand to fill the full interior volume of the shell. Wide zippers along vertical seams in the front of the device provide an easy access point for donning and doffing and permit the attachment of additional pneumatic pouches. A rock climbing style harness made of nylon webbing is worn at the hip and upper legs. It connects via side release buckles to the skirt to anchor it gently but securely to the body. We also developed a wheeled frame shown in Fig.~\ref{fig:components}A that can be added to the base of the FallGown for increased rigidity of the conical structure and reduction of sliding friction with the ground. When the frame is not used, a loop of soft webbing is attached in its place underneath the skirt to help reduce the tension applied to the bottom of the zippers.

\section{Evaluation}
\label{Section III}

The FallGown design was evaluated on its (A) stability and compressive load-bearing ability, (B) compactness and setup, (C) time and ease of donning and doffing, (D) burden on the wearer, (E) fall prevention, and (F) capability for usage monitoring. The tests described below were conducted by and with the research team only; readiness for safe testing and Institutional Review Board approval for use with human subjects, especially older adults with mobility challenges, are left for future work. With these tests, we aim to establish a methodology for evaluation of this approach to wearable support against several key assistive device needs identified by Morris et al.~\cite{Morris2023State}, demonstrate feasibility of the FallGown design, and set the stage for future comprehensive, ecologically valid user testing.

\subsection{Stability and Compressive Load}

\begin{figure}[h!]
    \centering
    \includegraphics[width=0.98\linewidth]{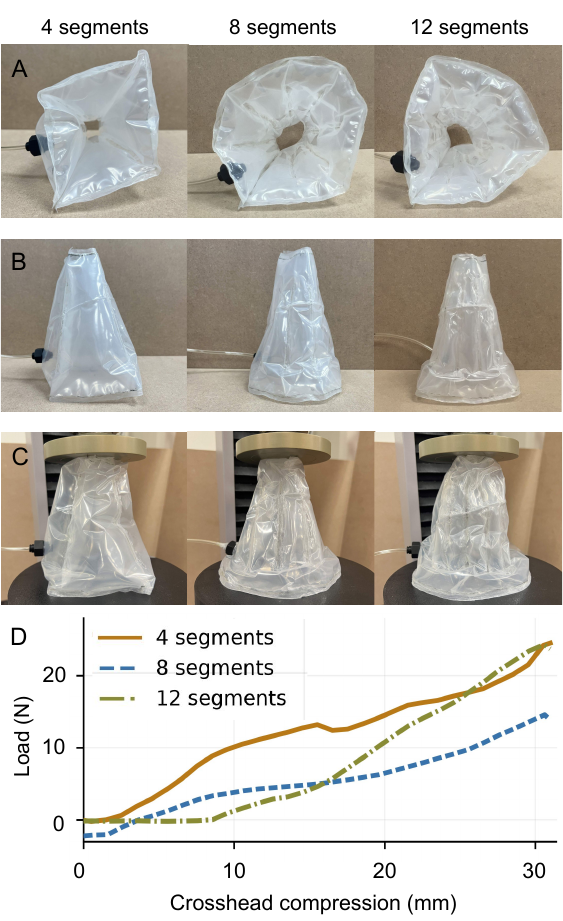}
    \caption{FallGown form factor compression tests performed with an MTS testing machine. (A) Bottom views, (B) side views in the uncompressed state, and (C) side views in the compressed state for the 4-segment, 8-segment, and 12-segment prototypes, from left to right. (D) Compressive force experienced by the prototypes as a function of compression distance.}
    \label{fig:loadtests}
    \vspace{-10mm}
\end{figure}

The FallGown uses a segmented conical structure to provide a snug fit at the user's waist and a broad contact area on the ground for stability. The inner spacers constrain the expansion of the device, preserving the central passage in which the user stands. Finer segmentation narrows each inflated wall, helping to reduce large inward and outward deformation. We fabricated scaled 4-, 8-, and 12-segment FallGown prototypes to examine both inflated geometry and vertical compression response. These scaled-down prototypes are shown in Fig.~\ref{fig:loadtests}. Each prototype was 155\,mm tall and inflated to approximately 0.1\,psi. As shown in Fig.~\ref{fig:loadtests}A, the more finely segmented prototypes had larger central openings at the base, providing the wearer with more walking room without increasing the size of the outer structure. However, we needed to verify that making the walls thinner did not cause the structure to become weaker under load.

To examine the prototypes' buckling response, compression tests were conducted in an MTS testing machine, with each prototype placed upright between two flat compression plates and loaded from the top down along its vertical axis. Tests were set up to compress each prototype vertically by 31\,mm (20\% of its nominal height) at 10\,mm/s. All three prototypes developed load continuously during compression, but their responses differed with segmentation, as shown in Fig.~\ref{fig:loadtests}D. The 4-segment prototype exhibited a less regular load response, including a local plateau and slight force reduction, indicating greater structural variation during compression. The 8-segment prototype showed a smoother and more continuous response, but remained noticeably nonlinear as the slope increased with compression. In contrast, the 12-segment prototype exhibited the smoothest and most nearly linear load development, especially in the second half of its loading cycle. None of the prototypes exhibited catastrophic buckling, indicating that the device will not collapse unexpectedly underneath a user's weight.

\subsection{Compactness and Setup}

\begin{figure}[b]
    \centering
    \includegraphics[width=0.98\linewidth]{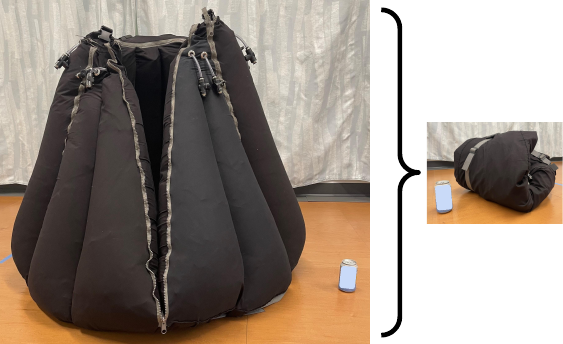}
    \caption{Size comparison between the device when fully inflated and fully deflated, with a 12-ounce (355\,mL) soda can (12\,cm tall) for comparison.}
    \label{fig:size}
\end{figure}

To evaluate the convenience of storing and setting up the device, we weighed it on a bathroom scale, packed it into various cardboard boxes, and measured the time it took to inflate each of the twelve pneumatic pouches from a vacuumed empty state to a fully inflated state. The skirt is 6.4\,kg, and the harness is 0.5\,kg. When deflated, the skirt can fit snugly into a 43$\times$41$\times$29\,cm cardboard box. The size difference between the inflated and uninflated states of the device is shown in Fig.~\ref{fig:size}. Using a 150\,L/min portable air pump (Motorenbau Electric Air Pump), it takes an average of $58\pm9$ seconds to inflate each of the twelve pouches to full pressure (0.2\,psi $\approx$ 1.4\,kPa). Including the time to switch the pump from pouch to pouch, the full setup takes about 15 minutes. The default pressure level was chosen to provide a comfortable level of support without risking the pneumatic connectors popping out, even during pressure spikes caused by user falls or collisions. With more robust permanent pneumatic fittings, higher pressure levels could be achieved for users desiring stiffer support, which would require a few extra seconds of inflation. Given the weight and size of the skirt, it is likely that an older adult user would need an assistant's help to set up the device, but it would not be a hassle to store it in a home or at a community center or retirement home.

\subsection{Donning and Doffing}

\begin{figure}[t]
    \centering
    \includegraphics[width=0.98\linewidth]{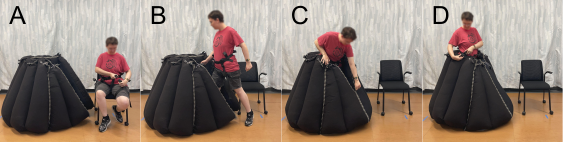}
    \caption{To don the FallGown, one must (A) put on the harness which goes around the waist and upper thighs, (B) step through the opening, (C) zip up and buckle the front closure, and (D) buckle the harness to the top of the skirt.}
    \label{fig:donning}
\end{figure}

To wear the FallGown, the user must put on the belt and leg loops of the harness, step into the skirt, zip up the skirt, and clip together the five buckles, as illustrated in Fig.~\ref{fig:donning}. From a starting position in which the harness is laid out on the ground in front of a user seated next to the inflated, unzipped skirt, we measured the time it took for a healthy young adult to don the device and then doff it. We repeated each trial 3 times. It takes an average of $89\pm33$ seconds for an able-bodied person to don the harness and skirt alone, and an average of $12\pm1$ seconds to doff it. For a helper, it takes an average of $145\pm26$ seconds to put the harness and skirt on a passive user and an average of $27\pm4$ seconds to take them off. The large variation in donning time is mainly due to the zipper sometimes getting stuck on the nearby fabric. The short donning time improves the likelihood that the user will maintain enthusiasm and energy levels for the ensuing exercise compared to a device that is difficult or slow to put on. The fast doffing time means a user can quickly exit the device if they get tired, need to use the restroom, or decide they are done exercising. We recognize that the process will likely take longer for our target population.

\subsection{Burden}

\begin{figure}[b]
    \centering
    \includegraphics[width=0.75\linewidth]{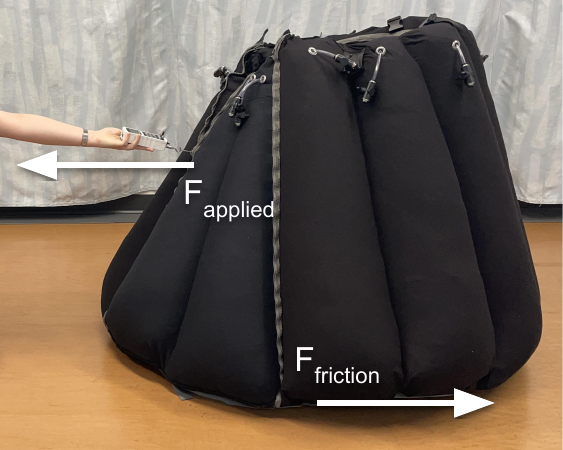}
    \caption{The drag force (burden) measurement setup, in which a force gauge is used to pull the skirt at a constant speed against friction horizontally across a floor.}
    \label{fig:drag}
\end{figure}

\begin{table}[t]
\caption{FallGown Movement Burden}
\label{table_burden}
\begin{center}
\begin{tabular}{|p{2cm}||p{1cm}|p{1cm}|p{1cm}|p{1cm}|}
\hline
Floor Type & Wood & Tiles & Carpet & Wheels on Tile \\
\hline
Drag Force (N) & 20.8 & 18.6 & 29.0 & 7.24 \\
\hline
\% of Body Weight & 2.59 & 2.32 & 3.61 & 0.90 \\
\hline
Metabolic Rate \% Increase & 30.3 & 27.0 & 43.2 & 11.1 \\
\hline
\end{tabular}
\end{center}
\end{table}

To measure the impact of the device on the effort required to walk, we dragged the skirt across several types of flooring. We pulled horizontally using a Mark-10 force gauge with a hook attachment, which we looped into the zipper pull tab about halfway up the skirt, as shown in Fig.~\ref{fig:drag}. We averaged the force during sliding motion when the skirt was moving at a steady pace to represent the burden experienced by a user in constant motion. The results of this test can be found in Table~I. To calculate the burden on the wearer, we approximated the average body weight of an individual over age 60 to be about 81.7\,kg \cite{CenterforHealthStatistics} to represent our intended user, and we constructed a polynomial approximation of the percent increase in metabolic rate as a function of the load applied at the waist as a percentage of body weight \cite{Gottschall2003}. Based on the drag friction, the wheeled frame is best suited for users who wish to frequently move laterally, while the skirt alone may work well for more stationary exercise activities.

\subsection{Fall Prevention}
\begin{figure}[t]
    \centering
    \includegraphics[width=0.98\linewidth]{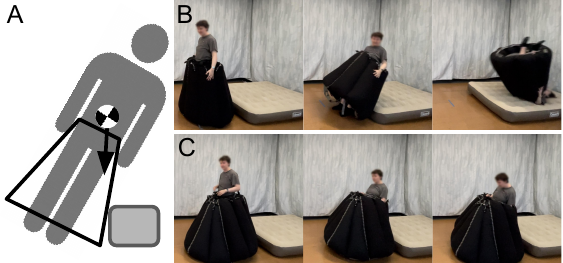}
    \caption{(A) The simulated fall setup, where the user's center of mass tips over the edge of the skirt and down to the mattress. (B) A backward fall in the 8-segment skirt, and (C) A prevented backward fall in the 12-segment skirt.}
    \label{fig:falltest}
\end{figure}

To test the device's ability to prevent falls, an experimenter wore the device and slowly leaned their body in various directions until they fell. An air mattress placed on the ground next to the skirt provided a simulated tripping hazard and a safe place to land. The setup of this experiment is illustrated in Fig.~\ref{fig:falltest}A. Three variations in the number of inflated pouches comprising the skirt were tested: 8 segments (103.5\,cm wide), 10 segments (125.5\,cm wide), and 12 segments (139.5\,cm wide). These versions all had the same waist measurement, but the added pouches caused variation in the steepness and the base perimeter of the device. The 12-segment version was tested alone and with the PVC frame shown in Fig.~\ref{fig:components}A. The experimenter tested falls in the forward, sideways, and backward directions. Each combination of skirt variation and fall direction was tested at least 3 times, with more repetitions when there were mixed results. The results of these tests are shown in Table~\ref{table_falls}.

The 8-segment version of the skirt was too narrow to prevent falls in any direction for the experimenter, resulting in falls like the one shown in Fig.~\ref{fig:falltest}B. The 10-segment version had mixed success; it prevented falls in the side and back only when the experimenter bent their knees (thus lowering their center of mass) while falling. The 12-segment design experienced the same issue; if the experimenter kept their body rigid, they tipped over, but if they relaxed their body, the fall was prevented, as shown in Fig.~\ref{fig:falltest}C. The 12-segment design prevented all falls in the forward direction because the asymmetry of the front panels made the cone slightly wider in the front. This difference indicates that three modifications should be made to the design: further widening the base, making it symmetrical, and raising the top to cover the body's center of mass. Adding the rigid frame implements these changes by stiffening and raising the base, which results in a higher success rate seen in the bottom rows of Table~\ref{table_falls}. The one sideways failure with the frame occurred when one of the wheels came off the frame, causing the device to tilt. With more robust wheels, success in all fall directions would be expected with the frame. However, this added stability comes at the cost of increased difficulty passing through doorways or narrow spaces because the frame prevents compression. 

Further testing would be required to determine whether the rigidity or the added height of the frame contributed more to stability. The skirt is 101\,cm tall, and the experimenter's center of mass is around 100\,cm based on their height \cite{Virmavirta2014}, so a 10\,cm height increase from the frame would have caused the skirt to more consistently cover their center of mass. Similar results might also be possible to achieve with a high-pressure ring-shaped pneumatic pouch around the base of the device to keep the cone more round and make it taller.

\begin{table}[t]
\caption{Fall Prevention Tests}
\label{table_falls}
\begin{center}
\begin{tabular}{|c|c|c|c|}
\hline
Number & Base Diameter & Fall Direction & Success Rate\\
of Segments & (cm) &   &   \\
\hline
8 & 103.5 & Front & \cellcolor[HTML]{f8b8d0} 0/3 \\
\hline
8 & 103.5 & Side & \cellcolor[HTML]{f8b8d0} 0/3 \\
\hline
8 & 103.5 & Back & \cellcolor[HTML]{f8b8d0} 0/3 \\
\hline
10 & 125.5 & Front & \cellcolor[HTML]{f8b8d0} 0/3 \\
\hline
10 & 125.5 & Side & \cellcolor[HTML]{d9e4ff} 2/4 \\
\hline
10 & 125.5 & Back & \cellcolor[HTML]{d9e4ff} 2/3 \\
\hline
12 & 139.5 & Front & \cellcolor[HTML]{b9e192} 3/3 \\
\hline
12 & 139.5 & Side & \cellcolor[HTML]{d9e4ff} 3/5 \\
\hline
12 & 139.5 & Back & \cellcolor[HTML]{d9e4ff} 3/6 \\
\hline
12 + Frame & 139.5 & Front & \cellcolor[HTML]{b9e192} 3/3 \\
\hline
12 + Frame & 139.5 & Side & \cellcolor[HTML]{d9e4ff} 7/8 \\
\hline
12 + Frame & 139.5 & Back & \cellcolor[HTML]{b9e192} 4/4 \\
\hline
\end{tabular}
\end{center}
\end{table}

\subsection{Usage Monitoring}

\begin{figure}[t]
    \centering
    \includegraphics[width=0.98\linewidth]{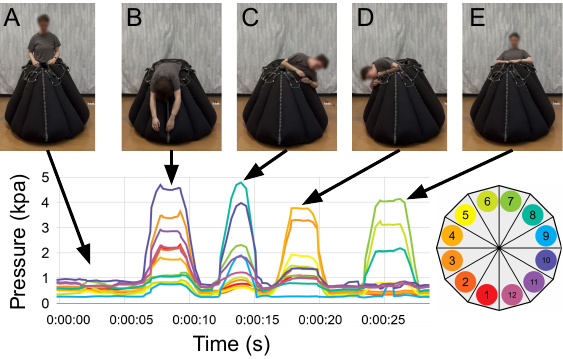}
    \caption{Pressure measured in the twelve pneumatic pouches during (A) walking, (B) leaning forward, (C) leaning to the user's left, (D) leaning to the user's right, and (E) leaning backward. The seam between segments 1 and 12 is at the front center of the FallGown.}
    \label{fig:pressures}
\end{figure}

To monitor the device and enable future features such as exercise guidance and responsive support, we attached air pressure sensors to each of the twelve pouches. We used HX710B Air Pressure Sensor Modules connected to an Arduino Nano 33 BLE to record the pressure values at approximately 4\,Hz. This microcontroller can stream the data wirelessly for applications requiring untethered movement. We tested the sensors by walking around in the device and then leaning/falling onto it. The pressure data was smoothed with a 3-point central moving average to reduce noise and plotted in Fig.~\ref{fig:pressures}. Peaks in the graph show when the user leans onto the device in different directions. In a lean, the user's weight compresses the pouches, resulting in increased internal pressure. The pouches that experienced the largest increase in pressure corresponded to the direction of the lean. In most of the leans, the user is supported primarily by one or two pouches due to their width relative to the experimenter's body. The forward lean (Fig.~\ref{fig:pressures}B) involves the most pouches because pouches 2, 1, 12, and 11 have narrower tops than the rest of the pouches to ensure a snug fit at the waist. 

On-board pressure sensing offers the opportunity for many extra device capabilities outside of fall prevention. By analyzing these peaks in real time, the device could sense leans that could lead to falls and issue alerts if the user does not recover within a specified time frame. The pressure data could also be used to analyze a user's balance and inform them if they tend to lean excessively in a certain direction so that they can correct their posture. Paired with an exercise routine, the device could check if the user is accurately following purposeful leaning stretches or dance moves. Long-term data collection could enable measurement of how a user's balance or need for support improves or worsens over time.

\section{Conclusions and Future Work}
The FallGown is stable under load, compact in storage, and quick to set up. Its burden upon the wearer depends on its configuration and the floor upon which it is used. Wider versions of the device prevent falls more frequently, but are less usable in confined environments. Sensors integrated into the device provide insight into users' posture and stability.

There are many changes that could be made to the FallGown's hardware. The sewing pattern should be modified to further widen the base and ensure more consistent prevention of falls. The hard plastic valves should be replaced with soft check valves like those used in beach balls, and the pressure sensors should be embedded into the pouches to keep the outer surface of the device smooth, soft, and comfortable. For versions of the device with a frame, the frame should be made more robust and collapsible for quick assembly. The frame could also be upgraded to have height adjustment to fit a wider range of user heights. For versions without a frame, low-friction materials and alternative geometries for the base should be explored to reduce drag friction, and additional ring pouches could be added to provide height adjustments. Patterned fabric could be used to increase the visual appeal of the device and permit user customization. Users could choose between designs ranging from frilly and feminine to minimalist and modern corresponding to their preferred self-expression through fashion. For users who, for reasons of taste or gender expression, do not want the device to look like a skirt, transparent versions made from clear PVC sheets or textile displays (i.e., luminous textiles) could be explored.

Future research may focus on the opportunities provided by the on-board pressure sensors. Passively, the device could monitor user posture and alert a user or caretaker in the case of a fall. With active pressure control, the device could respond to fall events by increasing pressure in the loaded pouches to help push a user back into a stable upright position. An actively pressurized device could provide nudges to guide a user during dances or exercises. The device could also be adjusted to provide differing levels of support depending on the user's daily energy level and health.

Future testing should also be conducted with the target population of older adults who struggle with weakness and balance. In speaking with older adults in our community about this project, we heard a lot of enthusiasm and curiosity for the FallGown, with many people interested in trying it for themselves. We also talked with clinicians about potential applications of the device for other populations with impaired balance, such as traumatic brain injury patients. Testing with these populations would prove the utility of the device and illuminate areas for further improvement.

\section{Acknowledgments}
The authors thank Izumi Yaskawa, the community members and staff at Live Oak Adult Day Services, and Durga Ghosh for valuable feedback, Alison Cao, Shujiro Shobayashi, and Stanley Wang for helping run the experiments, and Ava Chen for video editing assistance. OpenAI GPT-5.6 Terra was used in the writing and debugging of the simple Arduino pressure sensor code used in Section III.F, but not used to process the resulting data.



\bibliographystyle{IEEEtran} 
\bibliography{references} 
\end{document}